\documentclass[letterpaper, 10 pt, conference]{ieeeconf}

\IEEEoverridecommandlockouts
\usepackage{amsmath}
\usepackage{mathtools}
\usepackage{amssymb}
\usepackage{amsfonts}   
\usepackage{dsfont}

\usepackage[pdftex]{graphicx}
\usepackage{wrapfig}
\usepackage{capt-of}

\usepackage{booktabs}   
\usepackage{makecell}

\usepackage{algorithm}
\usepackage{algorithmic}

\usepackage[table, dvipsnames]{xcolor}  
\usepackage[table,dvipsnames]{xcolor}
\usepackage{cite}
\usepackage{hyperref}

\let\citep\cite
\usepackage[]{mdframed}

\usepackage{cleveref}

\usepackage{xparse}     
\usepackage{expl3}      
\usepackage{xspace}
\usepackage{lipsum}     
\usepackage{balance}

\definecolor{linkpink}{RGB}{237,0,140}

\crefname{section}{Sec.}{Secs.}
\crefname{figure}{Fig.}{Figs.}
\crefname{table}{Tab.}{Tabs.}
\crefname{equation}{Eq.}{Eqs.}

\DeclarePairedDelimiterX{\infdivx}[2]{(}{)}{%
  #1\;\delimsize\|\;#2%
}

\ExplSyntaxOn
\newcommand\latinabbrev[1]{
  \peek_meaning:NTF . {%
    \textit{#1}\@}%
  { \peek_catcode:NTF a {%
      \textit{#1}.\@ }%
    {\textit{#1}.\@}}}
\ExplSyntaxOff

\def\ie{\latinabbrev{i.e}}

\def\MethodName{Kinematic Interaction Transfer across Embodiments\xspace}
\def\MethodAcronym{KITE\xspace}

\newcommand{\src}{\mathrm{src}}
\newcommand{\tgt}{\mathrm{tgt}}

\title{\LARGE \bf
KITE: Decoupling Kinematics and Interaction for \\ Zero-Shot Cross-Embodiment Manipulation
}

\author{Qianxu Wang$^{1}$, Shengmiao Samuel Jin$^{1}$, Kuan Fang$^{1}$%
\thanks{$^{1}$All authors are with Cornell University, Ithaca, NY, USA.}%
}
\hypersetup{
    pdftitle={KITE: Decoupling Kinematics and Interaction for Zero-Shot Cross-Embodiment Manipulation},
    pdfauthor={Qianxu Wang, Shengmiao Samuel Jin, Kuan Fang}
}

\makeatletter
\newcommand{\compactearlyfigurecaptions}{%
    \setlength{\abovecaptionskip}{3pt}%
    \long\def\@makecaption##1##2{%
        \par\nointerlineskip
        \vskip\abovecaptionskip
        \noindent\parbox[t]{\hsize}{\normalfont\footnotesize\noindent ##1.~~##2}\par}}
\makeatother

\begin{document}
\maketitle
\thispagestyle{empty}
\pagestyle{empty}

\vspace{-15pt}


\begin{abstract}
Generalizing manipulation policies across robot embodiments remains difficult because standard policies tie the interactions needed for a task to the actions of a particular robot. We study zero-shot cross-embodiment manipulation, where a policy trained on source embodiments must be deployed on a structurally different target embodiment without additional task demonstrations. We introduce \textbf{K}inematic \textbf{I}nteraction \textbf{T}ransfer across \textbf{E}mbodiments (\textbf{KITE}), which decouples manipulation into embodiment-agnostic interaction prediction and embodiment-specific kinematic decoding, connected through a learned representation of interaction intent based on contact patterns. An interaction policy is trained to predict the intent using demonstration data only collected on a source robot, and shared across morphologically different target robots. A kinematic decoder realizes these intents by learning from synthetic data generated from each embodiment's kinematic model.
Through such decoupling, KITE can robustly transfer learned skills to a new embodiment given only its kinematic model, without recollecting any task-specific demonstration data.
We evaluate KITE on four manipulation tasks in simulation and the real world spanning transfer between parallel grippers, dexterous hands, and composite embodiments. KITE successfully achieves zero-shot transfer to structurally different target embodiments, and consistently outperforms state-of-the-art baselines.
Project website: \url{https://kite-manip.github.io/}.
\end{abstract}


\suppressfloats[t]
\begin{figure}[t]
    \centering
    \compactearlyfigurecaptions
    \noindent\includegraphics[width=\columnwidth]{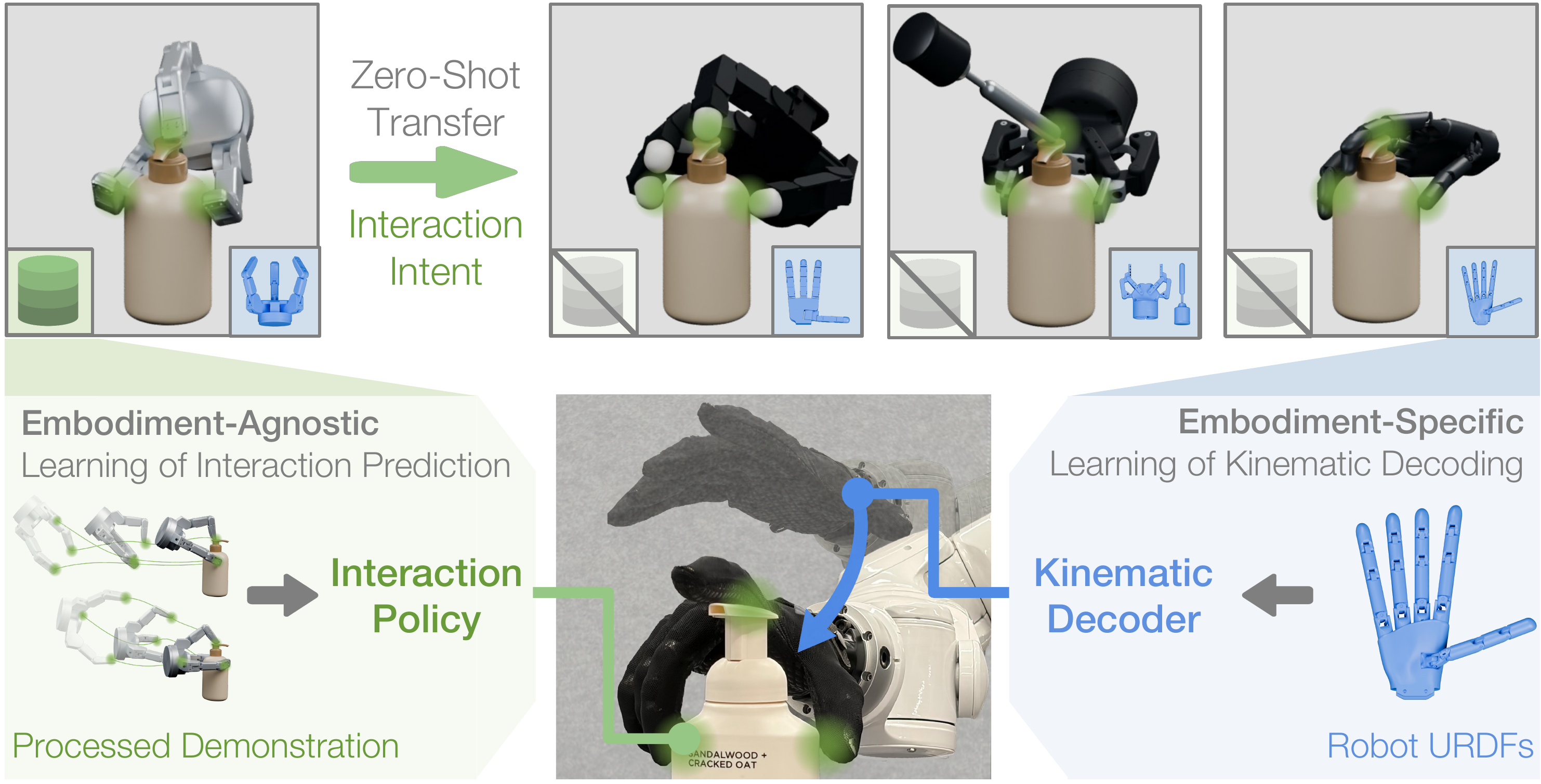}\par
    \caption{
    \MethodAcronym enables cross-embodiment transfer without needing to recollect demonstration data for target robotic manipulators.
    It decouples manipulation into embodiment-agnostic interaction prediction (\textcolor[HTML]{96C784}{green}) and embodiment-specific kinematic decoding (\textcolor[HTML]{598DE1}{blue}), connected by a learned latent representation of interaction intent.
    The former is implemented by an interaction policy trained on source demonstrations preprocessed into latent intents, while the latter is implemented by a kinematic decoder trained on synthetic data generated from each embodiment's kinematic model alone.
}
  \label{fig:teaser}
  \vspace{-2pt}
\end{figure}


\section{Introduction}

Manipulation policies nowadays remain bound to the embodiment on which the data was collected.
Their actions presuppose a particular kinematic structure, geometry, and control interface.
Consequently, task knowledge acquired on one robot rarely carries over to another, and adapting a learned skill to a new embodiment typically requires a new round of data collection.
The cost of robot learning thus grows with the number of embodiments as well as the number of tasks.
Cross-embodiment transfer aims to address this by reusing task knowledge from one robot on another~\cite{openx2023rtx}.
Its most ideal and demanding form is zero-shot transfer, in which task demonstrations are provided only for the source robot but not the target. A viable solution to this problem would change how robot learning scales across morphologically different hardware.
 
Existing approaches to cross-embodiment transfer for manipulation fall into three families, distinguished by where they place the burden of adapting to a new embodiment.
Generalist policies trained on multi-embodiment datasets~\citep{octo2024,doshi2024crossformer,liu2024rdt} pool demonstrations from many robots under a normalized action space.
Embodiment-conditioned policies~\citep{devin2017modular,huang2020smp,sferrazza2025body,patel2025getzero} expose the morphology to the network, but still learn the meaning of each action from demonstrations or reinforcement learning on the training embodiments.
Both families learn what to do and how a particular embodiment moves jointly from task demonstrations and task rollouts, so an embodiment without demonstrations lies outside the training distribution by construction.
Retargeting methods~\citep{handa2020dexpilot,qin2023anyteleop,pan2025spider} instead map source motion to the target through hand-specified part correspondences or a shared end-effector pose.
Such mappings require domain expertise for each pair of embodiments and become brittle when the interaction involves several contacts or a part with no counterpart on the target.
In each case, the knowledge of how a particular embodiment realizes an interaction is either purchased with demonstrations or written by hand.

In this paper, we propose to follow a different insight: what should transfer across embodiments is the robot-object interaction, rather than the robot actions that produce it. The success of a manipulation task is fundamentally determined by the interaction, \ie, how the relevant parts of the robot approach, contact, and manipulate the object. The knowledge of feasible interactions for a task can be inferred from successful trajectories on a source robot, and remains largely shareable across embodiments. On the other hand, given the intended interaction, how a robot should move to realize it is an embodiment-specific problem, determined mainly by the kinematic structure and geometry of the robot. Once the prediction and the realization of interactions are decoupled, the former can be learned from trajectories collected on the source robot, and the latter from the kinematic models of the source and target robots alone.

We present \textbf{\MethodName}~(\textbf{\MethodAcronym}), which enables zero-shot manipulation transfer through a contact-based decoupling of interaction prediction and realization.
As shown in \Cref{fig:teaser}, \MethodAcronym models the \emph{interaction intent} using a learned representation that describes how the object should be approached, contacted, and manipulated.
Around this intent, \MethodAcronym predicts actions through two decoupled stages: embodiment-agnostic interaction prediction and embodiment-specific kinematic decoding.
With demonstrations collected on the source robot, we train an \emph{interaction policy} to predict the interaction intent, which is shared across morphologically different robots.
A \emph{kinematic decoder} realizes the predicted intent through joint-space control at each time step. It is trained entirely on task-agnostic synthetic data from the robot's kinematic model, requiring no target demonstrations.
The intent representation and decoder bank, which contains one decoder per embodiment, are shared across tasks, so that a skill demonstrated on one embodiment executes on another by composing the policy with the decoder of the target.
To add an embodiment not included in the initial decoder training, we fine-tune only the decoder bank on synthetic kinematic data from both the existing and new embodiments. The intent space remains fixed, allowing existing interaction policies to be reused unchanged.
 
We evaluate \MethodAcronym on four manipulation tasks in both simulation and the real world.
Our simulation evaluation covers six robot embodiments, spanning a parallel gripper, dexterous hands, and a gripper paired with a separately actuated stick.
With one frozen decoder bank and one interaction policy per task, \MethodAcronym achieves an average score of $86.4\%$ across $12$ task--target settings in simulation, while the strongest baseline averages $53.3\%$ over its $10$ supported settings.
In the real world, the same decoder drives a Wuji hand from demonstrations given by a Robotiq gripper, an XHand, and a human hand.

\section{Related Work}


\subsection{Cross-embodiment policy learning}
Vision-language-action models use public robot datasets to achieve broad generalization within their training embodiments~\citep{kim2024openvla,black2024pi0}.
Additionally, models can train on heterogeneous multi-robot datasets using flexible observation/action spaces and unified action representations~\citep{openx2023rtx,octo2024,doshi2024crossformer,liu2024rdt}.
These approaches do not transfer zero-shot to morphologies outside their training distribution.
A complementary line explicitly incorporates embodiment structure: modular approaches factor robot- and task-specific components~\citep{devin2017modular,huang2020smp}, graph- and token-based architectures encode body structure~\citep{sferrazza2025body,patel2025getzero}, and canonical or learned latent action spaces align dexterous hands and multi-gripper families~\citep{wei2026onehand,yuan2025crossdex,jiang2026xlvla,bauer2025latentaction}.
These approaches improve transfer through modularity, structural conditioning, and action alignment, but do not generally separate task learning from the acquisition of embodiment-specific action semantics.
\MethodAcronym makes this separation explicit: an off-the-shelf imitation learner predicts shared contact-based interaction intent, while a kinematic decoder converts that intent into embodiment-specific commands.
A new embodiment is incorporated by learning its decoder from its kinematic model alone, without demonstrations on that embodiment.
Improvements in imitation learning can be adopted without changing this interface.

\subsection{Motion Retargeting and Correspondence-based Transfer}
Retargeting methods transfer demonstrations by matching source motion to a target robot through explicit correspondences and optimization~\citep{qin2023anyteleop,pan2025spider,mandi2025dexmachina}.
However, manually specified correspondences require expertise for each embodiment pair and can become inadequate for multi-contact interactions.
Per-frame or trajectory-level nonconvex optimization also introduces computational cost and initialization sensitivity~\citep{feng2026minimalist}.
\MethodAcronym instead learns a kinematic decoder that infers which robot part realizes each contact without manually specified correspondences, learns the mapping once from target kinematics for closed-loop policy execution, and replaces per-frame optimization with a single forward pass.

\subsection{Learning Contact-Based Dexterous Manipulation}
Prior work uses contact maps and geometric representations for dexterous grasp synthesis and cross-hand grasp transfer~\citep{li2023gendexgrasp,liu2023contactgen,wu2025cedex,wei2024drograsp,fei2025trograsp,shao2020unigrasp,attarian2023geomatch,xu2024manifoundation}, primarily addressing grasp configurations rather than sequential manipulation.
Object-centric representations, such as point tracks and flow, abstract away embodiment-specific actions but leave the manipulator's contact assignment unspecified~\citep{bharadhwaj2024track2act,xu2024flow,haldar2025pointpolicy}.
\MethodAcronym instead uses a contact-based latent intent that captures both interaction location and direction, providing a representation that is simultaneously rich enough to constrain embodiment-specific motor control and abstract enough for an embodiment-agnostic policy. 
This separates shared interaction intent from embodiment-specific motor control, enabling a learned policy to transfer to new morphologies without retraining.

\begin{figure*}[t]
    \vspace*{4pt}
    \centering
    \compactearlyfigurecaptions
    \noindent\includegraphics[width=\linewidth]{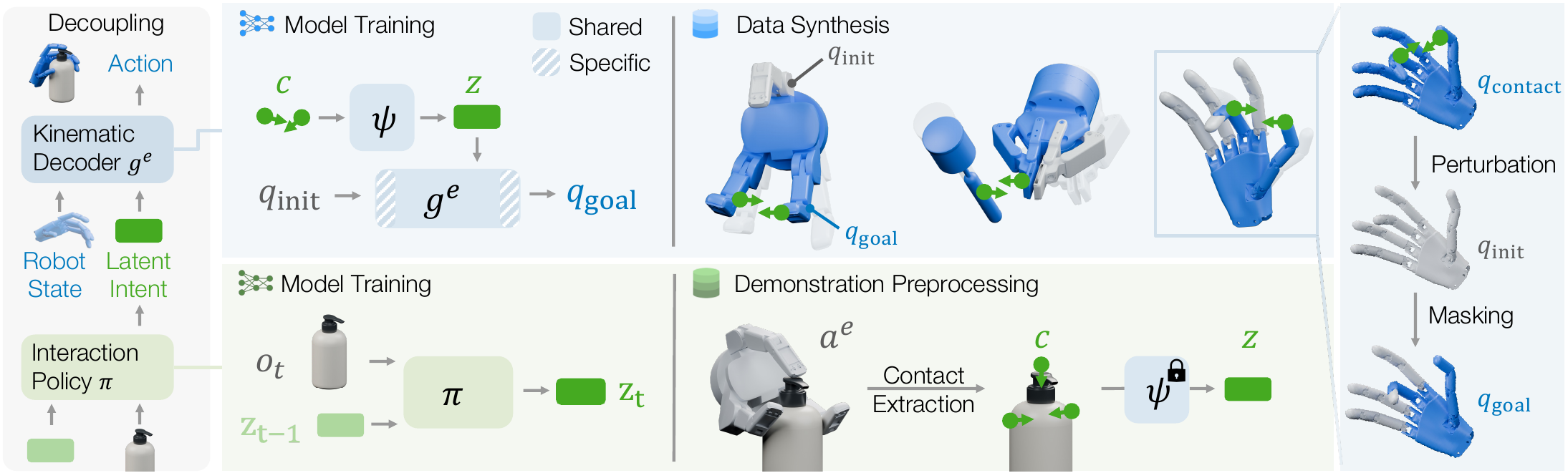}\par
    \caption{\textbf{Overview of \MethodAcronym.} We decouple manipulation into an embodiment-agnostic interaction policy $\pi$ and an embodiment-specific kinematic decoder $g^e$, connected by the latent intent $z$. Kinematic decoder $g^e$ learns from embodiments' kinematic models how that embodiment achieves an interaction intent; $\pi$ learns from source demonstrations preprocessed into latent intents.}
    \label{fig:method-overview}
    \vspace{-6pt}
\end{figure*}

\section{Problem Formulation}
\label{sec:problem}

We start by introducing a problem formulation of zero-shot cross-embodiment transfer for robotic manipulation based on limited expert demonstrations and known robot kinematic models. In this formulation, we separate the domains into the source ($\src$), embodiments for which demonstration data is provided during training, and the target ($\tgt$), embodiments on which we aim to evaluate the policy without demonstration data. They may differ in their kinematic structures, configuration spaces, and action spaces. The source dataset $\mathcal{D}^{\src}$ consists of expert demonstration trajectories, each a sequence of scene observations $o_t$ paired with source actions $a_t$. In principle, both domains can contain a set of multiple embodiments. Without loss of generality, we simplify the discussion to transferring from one source embodiment to one target embodiment for a single task.

For each embodiment $e \in \{\src, \tgt\}$, we assume a known kinematic model $\beta^e$, typically specified in Unified Robot Description Format (URDF), which determines its geometry at configuration $q^e \in \mathbb{R}^{d_e}$ through forward kinematics. Our goal is to learn a policy that solves the task on the target using $\mathcal{D}^{\src}$ and both kinematic models. We refer to this setting as \emph{zero-shot} since it \emph{does not use any task demonstration data on the target embodiment}.

\section{Method}

We present \MethodName~(\MethodAcronym), which enables zero-shot manipulation transfer through a contact-based decoupling of interaction prediction and realization (\Cref{sec:method-interface}).
In \emph{interaction prediction}, an embodiment-agnostic interaction policy $\pi$ predicts the next latent intent from the scene (\Cref{sec:method-ip}).
In \emph{kinematic decoding}, an embodiment-specific kinematic decoder $g^{e}$ produces a motor command that realizes this intent (\Cref{sec:method-kd}).
We describe decoder learning first because it defines the latent space in which the interaction policy is trained.
An overview is shown in \Cref{fig:method-overview}.

\subsection{Contact-Based Decoupling of Interaction and Kinematics}
\label{sec:method-interface}

The interface between the embodiment-agnostic interaction policy $\pi$ and the embodiment-specific kinematic decoder $g^{e}$ determines what information is passed between them. 
It must be expressive enough for manipulation while remaining embodiment-agnostic.
\emph{Contact} captures the robot-object interaction in the shared 3D workspace, abstracting over the embodiment that produces it. We therefore use a contact-based representation, \emph{interaction intent}, as this interface.

Formally, at each time step $t$, we define the interaction intent to be a sparse set
\begin{equation}
c_t = \{(x_{t,i},\, n_{t,i})\}_{i=1}^{K_t},
\label{eq:interaction-intent}
\end{equation}
where $x_{t,i}$ is a contact location and $n_{t,i}$ is the outward surface normal of the embodiment at that location, both expressed in the world frame. $c_t$ specifies the intended robot-object interaction without committing to any specific joint, link, fingertip, or end-effector that must realize it.

However, $c_t$ is unordered and has variable cardinality $K_t$, whereas standard policy architectures predict fixed-dimensional vectors. We therefore introduce an intent encoder $\psi$, a permutation-invariant set encoder shared by all embodiments, that maps each interaction intent to a fixed-dimensional \emph{latent intent}
\begin{equation}
z_t = \psi(c_t) \in \mathcal{Z}.
\label{eq:latent-intent}
\end{equation}
We train $\psi$ jointly with the kinematic decoders on data synthesized from the kinematic models (\Cref{sec:method-kd}).

\subsection{Embodiment-Specific Learning of Kinematic Decoding}
\label{sec:method-kd}

The kinematic decoder $g^{e}$ predicts a goal configuration of embodiment $e$ that realizes the engagement specified by the latent intent $z$, conditioned on its initial configuration, $q_{\mathrm{goal}}^{e} = g^{e}(z,\, q_{\mathrm{init}}^{e})$.
Here, $z = \psi(c)$ is derived from the interaction intent $c$ by the intent encoder $\psi$ (\Cref{sec:method-interface}).

We train the intent encoder $\psi$ and the decoder bank $\{g^{e}\}$ jointly on data synthesized from the kinematic models $\beta^{e}$ of the source and target embodiments, without any task demonstrations.
Since these components are reused across tasks, the encoder and decoder together must map the interaction intents each embodiment can produce to configurations that realize them.
This requires training triples $(c, q_{\mathrm{init}}, q_{\mathrm{goal}})$ that cover the embodiment's contact capabilities across its configuration space.
We synthesize these triples separately for each embodiment using the construction below, dropping the superscript $e$ where it is unambiguous.

Synthesizing intents with objects would require covering diverse object geometries, whereas every realizable intent can be described by points and outward normals on the embodiment's own surface at a certain configuration.
Sampling this surface across configuration space therefore lets us generate intents using forward kinematics alone, without any object model.
As shown in \Cref{fig:method-overview}, a fixed surface contact template $\mathcal{T}^{e}$ is precomputed once from $\beta^{e}$.
For each example, we sample a contact configuration $q_{\mathrm{contact}}$ within the embodiment's joint limits together with a subset $S$ of template indices, each identifying a surface point and its outward normal.
Sampled configurations are filtered for self-collision.
Forward kinematics at $q_{\mathrm{contact}}$ maps the selected points and their outward normals to the world frame, giving an interaction intent $c$ that $e$ realizes at $q_{\mathrm{contact}}$ by construction.
The initial configuration $q_{\mathrm{init}}$ is a Gaussian perturbation of $q_{\mathrm{contact}}$, with standard deviations of $5\%$ of each joint's range, $0.005$\,m for base position, and $5^\circ$ for base rotation, followed by clamping to the joint limits.

The remaining choice is how to supervise the goal configuration $q_{\mathrm{goal}}$.
The naive setting $q_{\mathrm{goal}} = q_{\mathrm{contact}}$ also supervises arbitrary motions of joints that do not affect the selected points, tying the latent intent to joint changes that $c$ does not specify.
We instead supervise contact-relevant joints and freeze the rest at $q_{\mathrm{init}}$,
\begin{equation}
q_{\mathrm{goal}} = m \odot q_{\mathrm{contact}} + (1 - m) \odot q_{\mathrm{init}},
\label{eq:goal-config}
\end{equation}
where $m \in \{0,1\}^{d_e}$ is the joint mask induced by $S$: $m_j = 1$ if joint $j$ affects a link containing a selected point, including upstream joints in the kinematic tree, and $m_j = 0$ otherwise.
The mask restricts the difference between $q_{\mathrm{goal}}$ and $q_{\mathrm{init}}$ to joints that affect the selected points.
We denote the resulting examples by $\mathcal{D}_{g}^{e}$.

The intent encoder $\psi$ is shared across embodiments, encouraging $\mathcal{Z}$ to carry consistent meaning across embodiments.
To improve parameter efficiency, the decoders $g^{e}$ share a common Transformer backbone; only the projectors that read $q_{\mathrm{init}}$ and write $q_{\mathrm{goal}}$ are embodiment-specific.
We train the intent encoder $\psi$ and the decoder bank $\{g^{e}\}$ jointly by supervised regression on the synthesized data,
\begin{equation}
\mathcal{L}_{g} = \sum_{e} \mathbb{E}_{(c,\, q_{\mathrm{init}},\, q_{\mathrm{goal}}) \sim \mathcal{D}_{g}^{e}} \left[ \left\| g^{e}(\psi(c),\, q_{\mathrm{init}}) - q_{\mathrm{goal}} \right\|_2^2 \right],
\label{eq:low-loss}
\end{equation}
where the outer sum runs over all training embodiments, including the target, as specified in \Cref{sec:problem}.

\subsection{Embodiment-agnostic Learning of Interaction Prediction}
\label{sec:method-ip}

After the intent encoder $\psi$ and the decoder bank $\{g^{e}\}$ are trained in \Cref{sec:method-kd}, training the \emph{interaction policy} $\pi$ reduces to a standard imitation-learning problem in $\mathcal{Z}$. In contrast to embodiment-conditioned policies %
that encode embodiment structure in custom architectures~\cite{sferrazza2025body,patel2025getzero}, we can use an off-the-shelf imitation learner and keep %
its architecture untouched. Given a base policy with a fixed-dimensional action space, we instantiate $\pi$ with only two changes: the action dimension is set to $\dim(\mathcal{Z})$, and the native proprioception input is replaced by the previous latent intent $z_{t-1}$, giving $z_t = \pi(o_t,\, z_{t-1})$. Feeding the source embodiment's joint state would reintroduce an embodiment-specific input into the interaction policy, whereas $z_{t-1}$ is defined in the shared latent space $\mathcal{Z}$. 

Training data for $\pi$ comes from source demonstrations preprocessed into latent intents. For each source trajectory, we select a fixed set of indices from the source surface template $\mathcal{T}^{\src}$. We score each index at every frame using the corresponding point's distance and normal alignment to the object~\citep{li2023gendexgrasp} and retain indices whose scores exceed a threshold in any frame. We cluster these candidates by their template positions, setting the cluster count from the average number of connected contact regions per frame, capped at six. From each cluster, we select the index with the largest sum of above-threshold scores across the trajectory, retaining at least two indices overall. Keeping these indices fixed, we apply forward kinematics at each source configuration $q_t^{\src}$ to their template positions and normals to form $c_t$ in the world frame, including frames without active contact. This gives the policy a motion target for the regions involved in the demonstrated interaction throughout the trajectory, including when they approach or withdraw from the object.

The frozen intent encoder from Eq.~\eqref{eq:latent-intent} converts each $c_t$ into a latent intent $z_t = \psi(c_t)$, giving training triples $(o_t, z_{t-1}, z_t)$. The policy sees only observations and latent intents, with no source joint states or actions provided to $\pi$. We train $\pi$ with the base policy's native loss on $z_t$, without any auxiliary objective.

\subsection{Deployment and Transfer}
\label{sec:method-deploy}

At deployment, the \emph{interaction policy} $\pi$ predicts the next latent intent from observation, and the embodiment-specific \emph{kinematic decoder} $g^{e}$ maps it to the next motor command,
\begin{equation}
a_t^{e} = g^{e}\big(\pi(o_t,\, z_{t-1}),\, q_t^{e}\big),
\label{eq:deploy}
\end{equation}
where the action is instantiated as the commanded next configuration $a_t^{e} := q_{t+1}^{e}$, so that $q_t^{e}$ and $a_t^{e}$ play the roles of $q_{\mathrm{init}}$ and $q_{\mathrm{goal}}$ of \Cref{sec:method-kd}. For target embodiments included in the pretrained decoder bank, the same $\pi$ is reused and only $g^{e}$ is replaced.


To support an embodiment outside the pretrained set, we add its input and output projectors and fine-tune the decoder bank on data synthesized from the kinematic models of both existing and new embodiments using Eq.~\eqref{eq:low-loss}. The intent encoder $\psi$ remains fixed, so existing interaction policies and preprocessed demonstrations can be reused unchanged.

\section{Experiments}
\label{sec:experiments}

\makeatletter
\newcommand{\compactexperimentcaptions}[1]{%
    \setlength{\abovecaptionskip}{3pt}%
    \long\def\@makecaption##1##2{%
        \par\nointerlineskip
        \vskip\abovecaptionskip
        \noindent\parbox[t]{\hsize}{\normalfont\footnotesize#1\noindent ##1.~~##2}\par}}
\makeatother

\setlength{\textfloatsep}{6pt minus 1pt}
\setlength{\floatsep}{6pt minus 1pt}
\setlength{\dbltextfloatsep}{7pt minus 1pt}

We address three questions through experiments and analysis.
\textbf{(Q1)} Can \MethodAcronym transfer manipulation skills across morphologically different embodiments, including parallel grippers, dexterous hands, and combinations of manipulators, on tasks with heterogeneous requirements, such as spatial precision and complex contact coordination?
\textbf{(Q2)} Can \MethodAcronym outperform learning-based cross-embodiment baselines and a trajectory retargeting baseline with access to source trajectories?
\textbf{(Q3)} How do \MethodAcronym's designs support the prediction and realization of interaction intents across embodiments?
We examine these questions in simulation (\Cref{sec:exp-setup}--\Cref{sec:exp-behavior}) and further deploy \MethodAcronym on a dexterous hand in the real world to test transfer from both robot and human demonstrations under real sensing and execution (\Cref{sec:exp-real}).

\subsection{Experimental Setup}
\label{sec:exp-setup}
To answer \textbf{Q1}, we evaluate four tasks across six robot embodiments, with human-hand demonstrations for real-world transfer.

\begin{figure}[t]
    \vspace*{5pt}
    \centering
    \compactexperimentcaptions{}
    \noindent\includegraphics[width=\columnwidth,pagebox=cropbox]{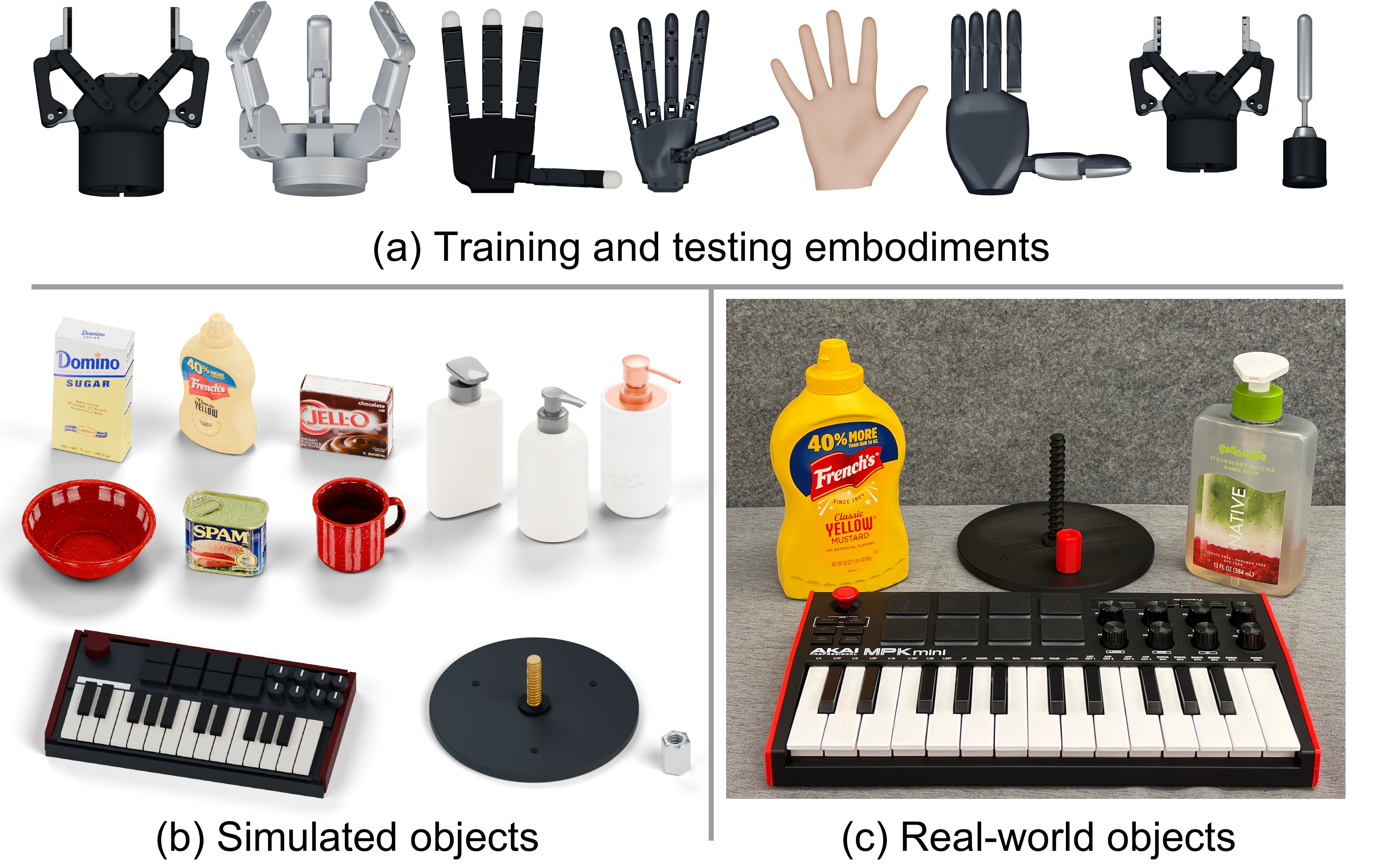}\par
    \caption{\textbf{Embodiments and objects used in the experiments.} (a) Six robot embodiments and the human hand. (b) Objects used for the four tasks in simulation. (c) Objects used for the four tasks in the real world.}
    \label{fig:experiment-scope}
    \vspace{-2pt}
\end{figure}

\noindent\textbf{Embodiments.} 
We use the Robotiq~2F-85 parallel gripper, three-finger Barrett, four-finger Allegro, and five-finger Wuji and XHand, as well as the human hand, shown in \Cref{fig:experiment-scope}a. Robotiq+Stick combines a gripper and a separately actuated stick to test whether simple manipulators can be coordinated to perform tasks demonstrated by a dexterous hand. Transfer from grippers and human hands to dexterous hands offers a way to use existing large-scale demonstration sources.

\noindent\textbf{Task design.}
We design four tasks with distinct demands, as shown in \Cref{fig:experiment-scope}b.
\emph{Objects Placing} transfers from Robotiq to Barrett, Allegro, and Wuji. Success requires picking up an object with randomized position and orientation and placing it inside a nearby bin. We use the same six objects in training and testing. This setting asks whether demonstrations from a simple gripper can teach a task to a dexterous hand.
\emph{Keyboard Pressing} transfers pressing three target keys from Wuji to Barrett, Allegro, and Robotiq+Stick, testing whether the interface preserves spatial precision; success requires depressing every target key past $50\%$ of its travel.
\emph{Bottles Pumping} uses the same transfer directions with three pump bottles. Success requires lifting the bottle at least $1$\,cm and depressing its pump at least $1$\,cm, testing coordination of supporting and pressing contacts. In particular, Robotiq+Stick tests whether contacts produced by one dexterous hand can be distributed across a gripper and a separate tool, extending transfer beyond changes in finger number.
\emph{Nut Rotating} transfers from XHand to Barrett, Allegro, and Wuji, requiring the hand to rotate a hex nut about a stud, which tests sustained contact coordination. Following the setup of Hsieh et al.~\citep{hsieh2026learning}, the nut is modeled as a revolute joint in simulation; the stud position and both source and target hand bases are fixed. For each episode, we divide the target's net rotation by the corresponding demonstration's net rotation and clip the ratio to $[0,1]$. We define success rate for this task as the mean ratio across episodes, expressed as a percentage.

\noindent\textbf{Training and evaluation protocols.}
The same intent encoder and decoder bank are pretrained using the kinematic models of all involved embodiments, then frozen and reused across all four tasks; only the task-specific interaction policies change. Target kinematics are available during pretraining, but no target-task demonstrations are used.

For objects placing, we synthesize $2{,}400$ source demonstrations by combining GraspGen-X grasp proposals~\citep{han2026graspgenx} with a rule-based policy. Keyboard pressing and bottles pumping each use $800$ scripted demonstrations. Nut rotating uses $800$ demonstrations from an RL-trained XHand policy following Hsieh et al.~\citep{hsieh2026learning}. All interaction policies use iDP3~\citep{ze2025generalizable} with object-only point clouds from two external cameras as scene observations. We evaluate each target on $100$ held-out episodes and report success rates for all four tasks. For the main comparison, we use three random seeds to train \MethodAcronym's interaction prediction policy and each learning-based baseline, and to run SPIDER's retargeting optimization. We keep the data splits and evaluation episodes fixed and report the mean and sample standard deviation (SD) across seeds.

\subsection{Zero-Shot Cross-Embodiment Transfer}
\label{sec:exp-transfer}

To answer \textbf{Q2}, we compare \MethodAcronym with two types of baselines: embodiment-conditioned methods and trajectory retargeting. The first type conditions action or pose prediction on robot structure; the second transfers source trajectories through optimization. Results are reported in \Cref{tab:experiment-results}.

\begin{table*}[t]
    \centering
    \vspace*{6pt}
    \setlength{\abovecaptionskip}{3pt}
    \makeatletter
    \long\def\@makecaption#1#2{%
        \parbox[t]{\hsize}{\normalfont\footnotesize\noindent #1.~~#2}\par
        \vskip\abovecaptionskip}
    \makeatother
    \caption{Cross-embodiment transfer success rates (\%, higher is better), reported as mean $\pm$ standard deviation across three random seeds ($100$ episodes per seed and target), except Oracle Intent, which shows success rates only. Avg. is the unweighted mean across applicable task--target settings, excluding N/A. For nut rotating, success rate is the mean clipped rotation ratio defined in \Cref{sec:exp-setup}. Upper block: decoder-only comparison using source trajectories. Lower block: closed-loop policies and single-pose adaptations of OHRA and D(R,O). Bold marks the best reported mean within each block. N/A denotes settings not covered by the evaluated adaptations; R+S: Robotiq+Stick.}
    \label{tab:experiment-results}
    \fontsize{7}{8.4}\selectfont
    \newcommand{\resultcell}[2]{%
        \mbox{#1{\fontsize{6}{7.2}\selectfont\hspace{0.4pt}\textpm\hspace{0.3pt}#2}}}
    \setlength{\tabcolsep}{0.5pt}
    \renewcommand{\arraystretch}{1.12}
    \begin{tabular*}{\textwidth}{@{\extracolsep{\fill}}lcccccccccccc@{\hspace{3pt}}c@{}}
        \toprule
        Task & \multicolumn{3}{c}{Objects Placing} & \multicolumn{3}{c}{Keyboard Pressing} & \multicolumn{3}{c}{Bottles Pumping} & \multicolumn{3}{c}{Nut Rotating} & \\
        \cmidrule(lr){2-4} \cmidrule(lr){5-7} \cmidrule(lr){8-10} \cmidrule(lr){11-13}
        Source & \multicolumn{3}{c}{Robotiq} & \multicolumn{3}{c}{Wuji} & \multicolumn{3}{c}{Wuji} & \multicolumn{3}{c}{XHand} & \\
        Target & Barrett & Allegro & Wuji & Barrett & Allegro & R+S & Barrett & Allegro & R+S & Barrett & Allegro & Wuji & Avg. \\
        \midrule
        \textit{SPIDER} & \resultcell{20.0}{0.0} & \resultcell{22.0}{1.0} & \resultcell{61.3}{3.2} & \resultcell{96.0}{0.0} & \resultcell{92.3}{0.6} & N/A & \resultcell{32.0}{0.0} & \resultcell{63.0}{0.0} & N/A & \resultcell{44.9}{3.3} & \resultcell{54.3}{2.2} & \resultcell{47.3}{1.0} & 53.3 \\
        \addlinespace[2pt]
        \MethodAcronym{} (Oracle Intent) & \textbf{82.0} & \textbf{85.0} & \textbf{85.0} & \textbf{99.0} & \textbf{97.0} & \textbf{100.0} & \textbf{88.0} & \textbf{89.0} & \textbf{99.0} & \textbf{64.4} & \textbf{85.6} & \textbf{89.6} & \textbf{88.6} \\
        \midrule
        \textit{GET-Zero} & \resultcell{9.0}{15.6} & \resultcell{0.0}{0.0} & \resultcell{1.0}{1.7} & \resultcell{0.0}{0.0} & \resultcell{17.7}{15.7} & N/A & \resultcell{0.0}{0.0} & \resultcell{0.0}{0.0} & N/A & \resultcell{0.0}{0.0} & \resultcell{9.9}{8.8} & \resultcell{6.3}{8.1} & 4.4 \\
        \addlinespace[2pt]
        \textit{OHRA} & N/A & N/A & N/A & \resultcell{0.0}{0.0} & \resultcell{0.0}{0.0} & N/A & N/A & N/A & N/A & N/A & N/A & N/A & 0.0 \\
        \addlinespace[2pt]
        \textit{D(R,O)} & \resultcell{23.3}{4.2} & \resultcell{3.3}{1.5} & \resultcell{5.3}{2.1} & \resultcell{0.0}{0.0} & \resultcell{0.0}{0.0} & N/A & N/A & N/A & N/A & N/A & N/A & N/A & 6.4 \\
        \addlinespace[2pt]
        \MethodAcronym{} (w/o Latent) & \resultcell{42.0}{3.0} & \resultcell{46.3}{2.1} & \resultcell{44.3}{3.5} & \resultcell{61.3}{4.9} & \resultcell{62.0}{1.0} & \resultcell{65.0}{6.2} & \resultcell{44.0}{1.0} & \resultcell{62.3}{4.9} & \resultcell{63.0}{0.0} & \resultcell{44.1}{1.4} & \resultcell{63.3}{1.9} & \resultcell{27.3}{0.6} & 52.1 \\
        \addlinespace[2pt]
        \MethodAcronym{} (Full) & \resultcell{\textbf{76.0}}{3.6} & \resultcell{\textbf{85.3}}{2.1} & \resultcell{\textbf{82.3}}{1.5} & \resultcell{\textbf{96.0}}{1.0} & \resultcell{\textbf{95.0}}{0.0} & \resultcell{\textbf{100.0}}{0.0} & \resultcell{\textbf{86.3}}{10.0} & \resultcell{\textbf{86.3}}{3.2} & \resultcell{\textbf{92.0}}{1.0} & \resultcell{\textbf{64.0}}{0.5} & \resultcell{\textbf{83.9}}{1.4} & \resultcell{\textbf{89.3}}{1.1} & \textbf{86.4} \\
        \bottomrule
    \end{tabular*}
\end{table*}

\begin{figure}[!t]
    \vspace*{5pt}
    \centering
    \compactexperimentcaptions{}
    \noindent\includegraphics[width=\columnwidth,pagebox=cropbox]{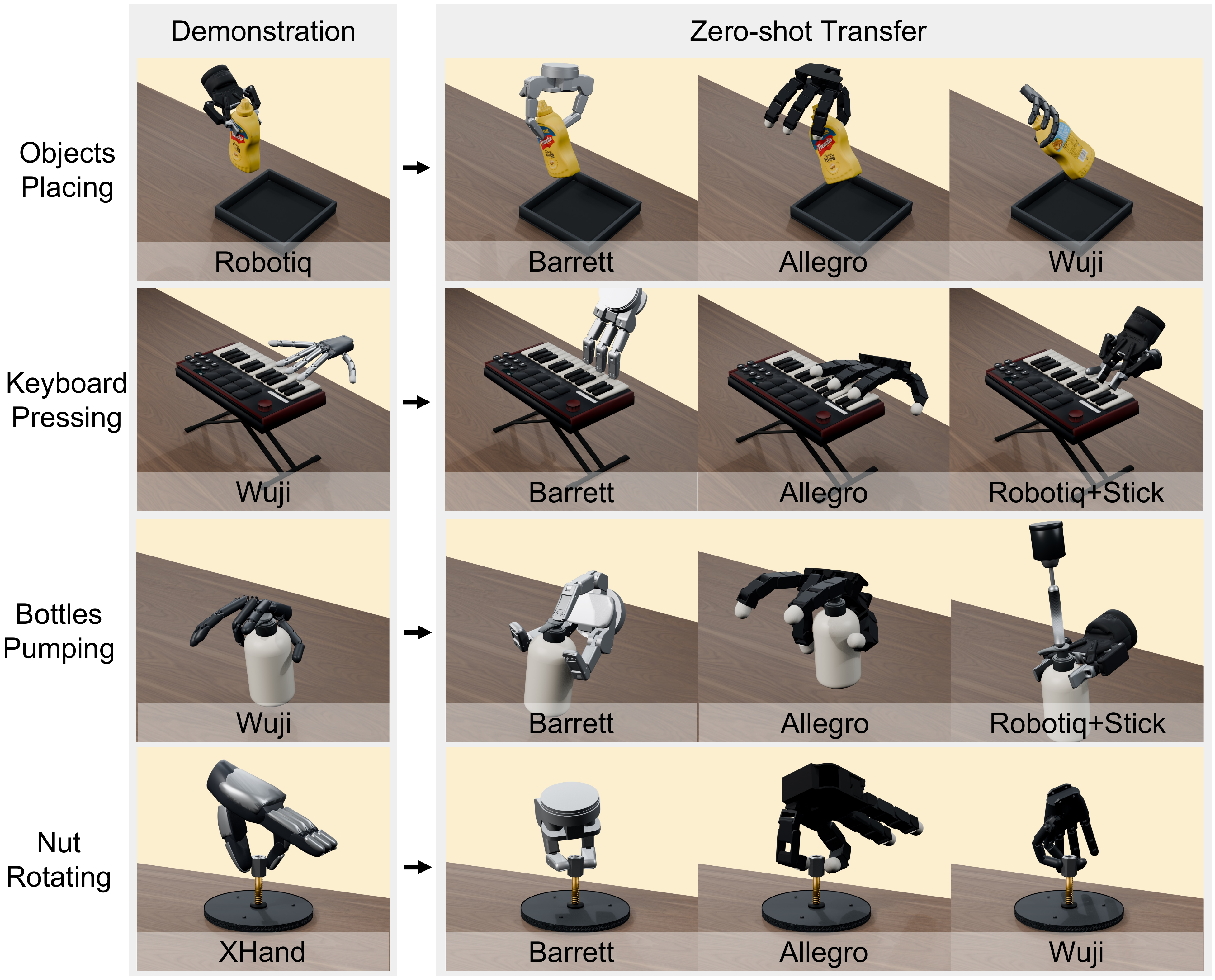}\par
    \caption{\textbf{Representative cross-embodiment transfer settings.} Source demonstrations and zero-shot target execution for objects placing, keyboard pressing, bottles pumping, and nut rotating.}
    \label{fig:experiment-transfer} 
    \vspace{-2pt}
\end{figure}

The comparison between \MethodAcronym{} and embodiment-conditioned methods suggests that learning to realize intended contacts can make more effective use of the target's kinematic model than conditioning predictions on its structure alone. \textit{GET-Zero}~\citep{patel2025getzero} predicts closed-loop joint actions using graph-conditioned attention and learns from embodiment-specific experts with an auxiliary forward-kinematics loss. Its reported transfer is across variants of the same hand design. With only one source embodiment per task here, the action mapping must generalize to different hand designs from source demonstrations alone; predicting forward kinematics does not directly supervise how target actions should realize the task's contacts. This can limit transfer despite explicit graph conditioning. \textit{OHRA}~\citep{wei2026onehand} and \textit{D(R,O)}~\citep{wei2024drograsp} predict target poses conditioned on embodiment structure. We train these adaptations on source demonstrations and combine a single predicted pose with scripted execution. Besides limited transfer ability to structurally different targets, they are further constrained by scripted execution, which cannot correct errors as the task progresses.

\begin{figure}[!t]
    \centering
    \compactexperimentcaptions{}
    \noindent\includegraphics[width=\columnwidth,pagebox=cropbox]{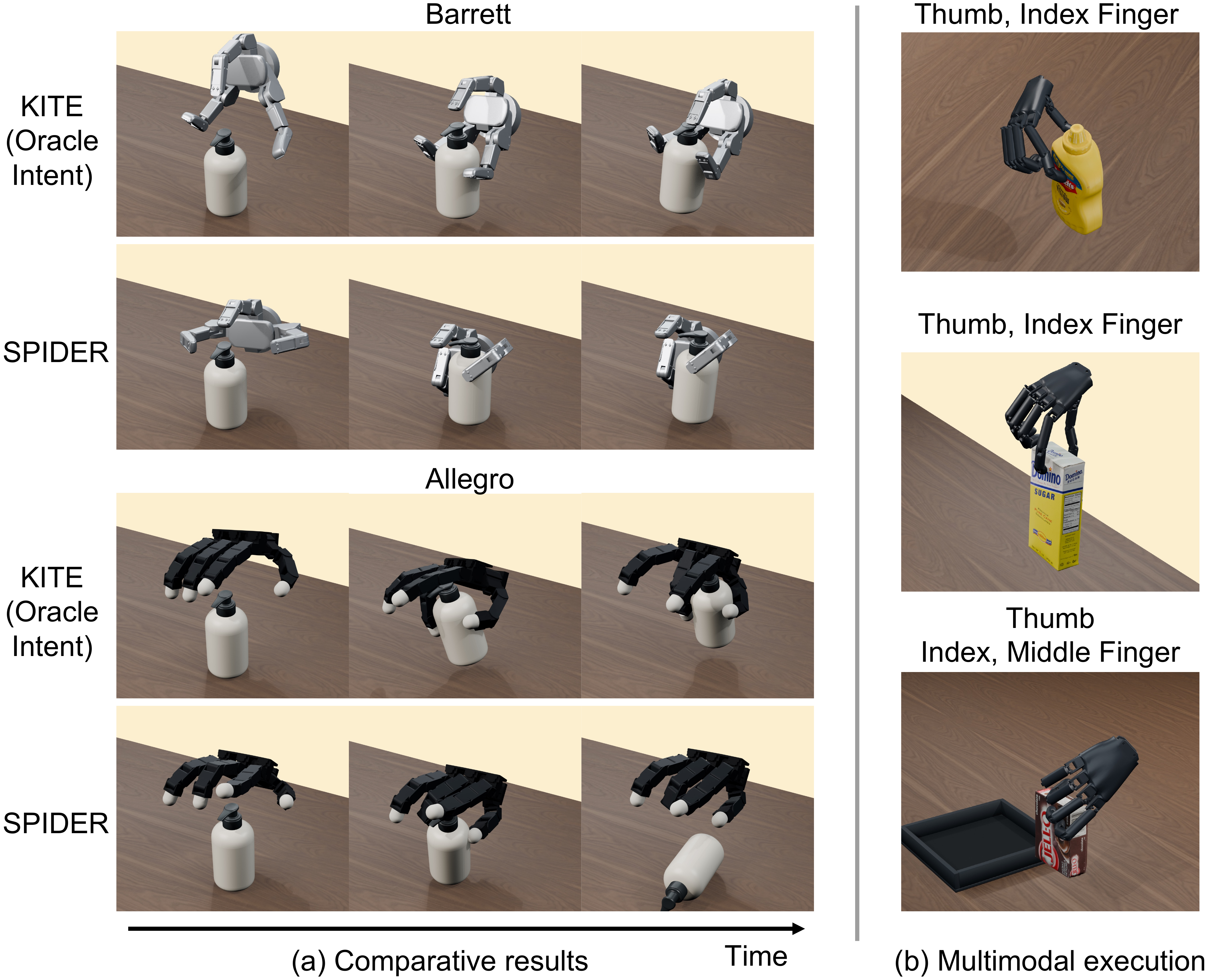}\par
    \caption{\textbf{(a) Decoder-only comparison.} Representative bottles pumping rollouts using source-demonstration trajectories. SPIDER loses contact during execution, whereas Oracle Intent maintains supporting and pressing contacts on Allegro and Barrett. \textbf{(b) Finger use across objects.} During objects placing, the kinematic decoder uses different finger combinations across objects to solve the task.}
    \label{fig:decoder-behavior}
    \vspace{-2pt}
\end{figure}

Trajectory retargeting methods such as \textit{SPIDER}~\citep{pan2025spider} have direct access to the source trajectory and object motion, whereas our full policy must predict interaction intents from live observations. To isolate kinematic decoding from interaction prediction, we compare SPIDER with \emph{\MethodAcronym{} (Oracle Intent)}, which supplies the frozen kinematic decoder with latent intents encoded from source demonstrations. SPIDER uses manually specified part correspondences and contact-based optimization to obtain target motion. Oracle Intent exceeds SPIDER in every reported comparison. SPIDER transfers keyboard pressing well, but the gap is larger on bottles pumping, where supporting and pressing contacts must be coordinated throughout execution. The rollouts in \Cref{fig:decoder-behavior}a show SPIDER losing contact while the kinematic decoder preserves the interaction. This supports learning how the target can realize the intended contacts, beyond specifying which target parts correspond to the source.

The N/A entries reflect limits of the evaluated baseline implementations: our GET-Zero adaptation excludes the independently controlled Robotiq+Stick composite; OHRA's palm-and-finger representation excludes the gripper source and composite target; the pose-generation adaptations do not cover Robotiq+Stick or the changing contact sequences of bottles pumping and nut rotating; and SPIDER lacks a prescribed correspondence for Robotiq+Stick. \MethodAcronym uses the same contact interface across these structures, supporting zero-shot transfer between fundamentally different embodiments on tasks with distinct contact requirements.

\subsection{Ablations and Analysis}
\label{sec:exp-ablations}
We compare \MethodAcronym (Full) with two variants  
in \Cref{tab:experiment-results} and separately examine decoder supervision.

\noindent\textbf{Latent intent interface.}
To test the benefit of the latent intent interface (\textbf{Q3}), we use \emph{\MethodAcronym{}~(w/o Latent)}, which predicts raw contact slots instead of the latent intent, retaining the frozen intent encoder and kinematic decoder for execution. Average success drops by over $30$ percentage points on each of the first three tasks, with the largest individual drop on Wuji nut rotating ($89.3\%$ to $27.3\%$). Raw contact sets are unordered and may vary in cardinality, while standard action heads predict fixed-dimensional vectors. This misalignment makes it difficult for the policy to learn the semantic meaning of the action space. 

\noindent\textbf{Closed-loop policy versus oracle intent.}
\emph{Oracle Intent} replaces predicted latent intents with those encoded from source demonstrations while keeping the kinematic decoder fixed. Full and Oracle Intent achieve broadly comparable performance; on nut rotating, their success rates differ by less than $2$ percentage points on all three targets. This supports the latent interface design, which allows a standard imitation learner to predict intents well enough to approach oracle performance.

\begin{figure*}[!t]
    \centering
    \vspace*{5pt}
    \compactexperimentcaptions{\raggedright}
    \makeatletter
    \long\def\@makecaption#1#2{%
        \par\nointerlineskip
        \ifx\@captype\@IEEEtablestring
            \noindent\parbox[t]{\hsize}{\normalfont\footnotesize\raggedright\noindent #1.~~#2}\par
            \vskip\abovecaptionskip
        \else
            \vskip\abovecaptionskip
            \noindent\parbox[t]{\hsize}{\normalfont\footnotesize\raggedright\noindent #1.~~#2}\par
        \fi}
    \makeatother
    \footnotesize
    \newsavebox{\decodercaptionpanel}
    \newsavebox{\extensionpanel}
    \newsavebox{\initialposepanel}
    \newsavebox{\realresultspanel}
    \newlength{\resultsidewidth}
    \newlength{\resultpanelgap}
    \setlength{\resultsidewidth}{\dimexpr(\linewidth-\columnsep-12pt)/4\relax}
    \setlength{\resultpanelgap}{6pt}
    \sbox{\decodercaptionpanel}{%
        \begin{minipage}{\resultsidewidth}
            \centering
            \caption{\textbf{Decoder supervision.} Mean $\pm$ standard deviation across three random seeds.}
            \label{fig:kinematics-sufficiency}
        \end{minipage}}
    \sbox{\extensionpanel}{%
        \begin{minipage}{\resultsidewidth}
            \centering
            \captionof{table}{Objects placing (\%, mean $\pm$ SD; three seeds). N/A: no Wuji decoder.\strut}
            \label{tab:new-embodiment}
            \fontsize{7}{8}\selectfont
            \newcommand{\extensionresultcell}[2]{%
                \mbox{#1{\fontsize{6}{7.2}\selectfont\hspace{0.4pt}\textpm\hspace{0.3pt}#2}}}
            \setlength{\tabcolsep}{1pt}
            \renewcommand{\arraystretch}{0.95}
            \begin{tabular*}{\linewidth}{@{\extracolsep{\fill}}lcc@{}}
                \toprule
                Target & \makecell{Joint training\\w/o Wuji} & \makecell{After adding\\Wuji} \\
                \midrule
                Barrett & \extensionresultcell{74.7}{3.1} & \extensionresultcell{74.0}{2.6} \\
                Allegro & \extensionresultcell{84.0}{2.6} & \extensionresultcell{84.3}{2.1} \\
                Wuji & N/A & \extensionresultcell{80.0}{3.6} \\
                \bottomrule
            \end{tabular*}
        \end{minipage}}
    \sbox{\initialposepanel}{%
        \begin{minipage}{\resultsidewidth}
            \centering
            \captionof{table}{Wuji objects placing under initial-pose perturbations.}
            \label{tab:initial-pose-robustness}
            \fontsize{7}{8}\selectfont
            \setlength{\tabcolsep}{1pt}
            \renewcommand{\arraystretch}{0.95}
            \begin{tabular*}{\linewidth}{@{\extracolsep{\fill}}lc@{}}
                \toprule
                Perturbation & Success (\%) \\
                \midrule
                Base (no perturbation) & 85 \\
                $5$\,cm, $30^\circ$ & 80 \\
                $20$\,cm, $60^\circ$ & 78 \\
                Flipped $180^\circ$ & 66 \\
                $30$\,cm shift & 57 \\
                \bottomrule
            \end{tabular*}
        \end{minipage}}
    \sbox{\realresultspanel}{%
        \begin{minipage}{\resultsidewidth}
            \centering
            \captionof{table}{Real-world transfer to Wuji ($10$ trials/task).\strut}
            \label{tab:real-world-results}
            \fontsize{7}{8}\selectfont
            \setlength{\tabcolsep}{1pt}
            \renewcommand{\arraystretch}{0.95}
            \begin{tabular*}{\linewidth}{@{\extracolsep{\fill}}l@{\hspace{2pt}}l@{\hspace{2pt}}c@{}}
                \toprule
                Task & Source & \makecell{Success\\(\%)} \\
                \midrule
                Objects Placing & Robotiq & 80 \\
                Keyboard Pressing & Human & 80 \\
                Bottles Pumping & Human & 60 \\
                Nut Rotating & XHand & 70 \\
                \bottomrule
            \end{tabular*}
        \end{minipage}}
    \noindent\hbox to\linewidth{%
        \vbox{\hbox{\usebox{\extensionpanel}}\kern0pt}%
        \kern\resultpanelgap
        \vbox{\hbox{\usebox{\initialposepanel}}\kern0pt}%
        \kern\columnsep
        \vbox{\hbox{\usebox{\realresultspanel}}\kern0pt}%
        \kern\resultpanelgap
        \vbox{\offinterlineskip
            \hbox{\includegraphics[width=\resultsidewidth]{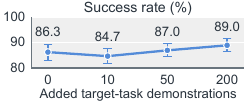}}%
            \hbox{\usebox{\decodercaptionpanel}}\kern-0.12pt}\hss}
\end{figure*}

\noindent\textbf{Adding a new embodiment.}
To test whether an existing policy can control a newly added embodiment, we exclude Wuji from intent encoder and kinematic decoder pretraining and train the objects-placing interaction policy on Robotiq demonstrations. Keeping the encoder and policy frozen, we add Wuji's synthetic kinematic data to the training set and fine-tune the decoder bank for $50$ epochs ($10$ hours on a single RTX~5090). \Cref{tab:new-embodiment} compares success before and after adding Wuji, using the same policy and evaluation episodes.
After fine-tuning, Wuji performs comparably to joint pretraining (\Cref{tab:experiment-results}), while performance on the existing embodiments remains similar.

\noindent\textbf{Kinematics-only decoder supervision.}
To test whether synthetic data from kinematic models alone suffice for decoder training, we add $10$, $50$, or $200$ target-task demonstrations in a separate Allegro bottles pumping diagnostic (\Cref{fig:kinematics-sufficiency}). The limited benefit from added demonstrations (at most $2.7$ percentage points) supports learning target-side contact realization from kinematics alone in this setting.

\subsection{Decoder Behavior and Initialization Sensitivity}
\label{sec:exp-behavior}

To further examine how the kinematic decoder realizes interaction intents (\textbf{Q3}), we inspect Wuji during objects placing. Although all demonstrations come from the same Robotiq gripper, the decoder can use different finger combinations across objects (\Cref{fig:decoder-behavior}b). 
Unlike retargeting methods that rely on fixed part correspondences, the decoder realizes these intents without a prescribed source-to-target mapping.

The kinematic decoder realizes each intent conditioned on the target's current configuration, so its ability to realize the interaction intents may depend on the initial pose. To test this sensitivity (\textbf{Q3}), we evaluate \emph{Oracle Intent} on Wuji objects placing, replaying the same source-intent sequences from different initial hand poses. In \Cref{tab:initial-pose-robustness}, success remains close to the $85\%$ base rate under a $5$\,cm, $30^\circ$ perturbation ($80\%$), but falls to $66\%$ with a $180^\circ$ flip and $57\%$ with a $30$\,cm shift. This is consistent with the decoder's local training construction (\Cref{sec:method-kd}), which learns to recover contact-relevant configurations from nearby perturbed poses but does not guarantee intent realization from arbitrary initializations.

\subsection{Real-World Deployment}
\label{sec:exp-real}

\begin{figure}[!t]
    \centering
    \compactexperimentcaptions{}
    \noindent\includegraphics[width=\columnwidth,pagebox=cropbox]{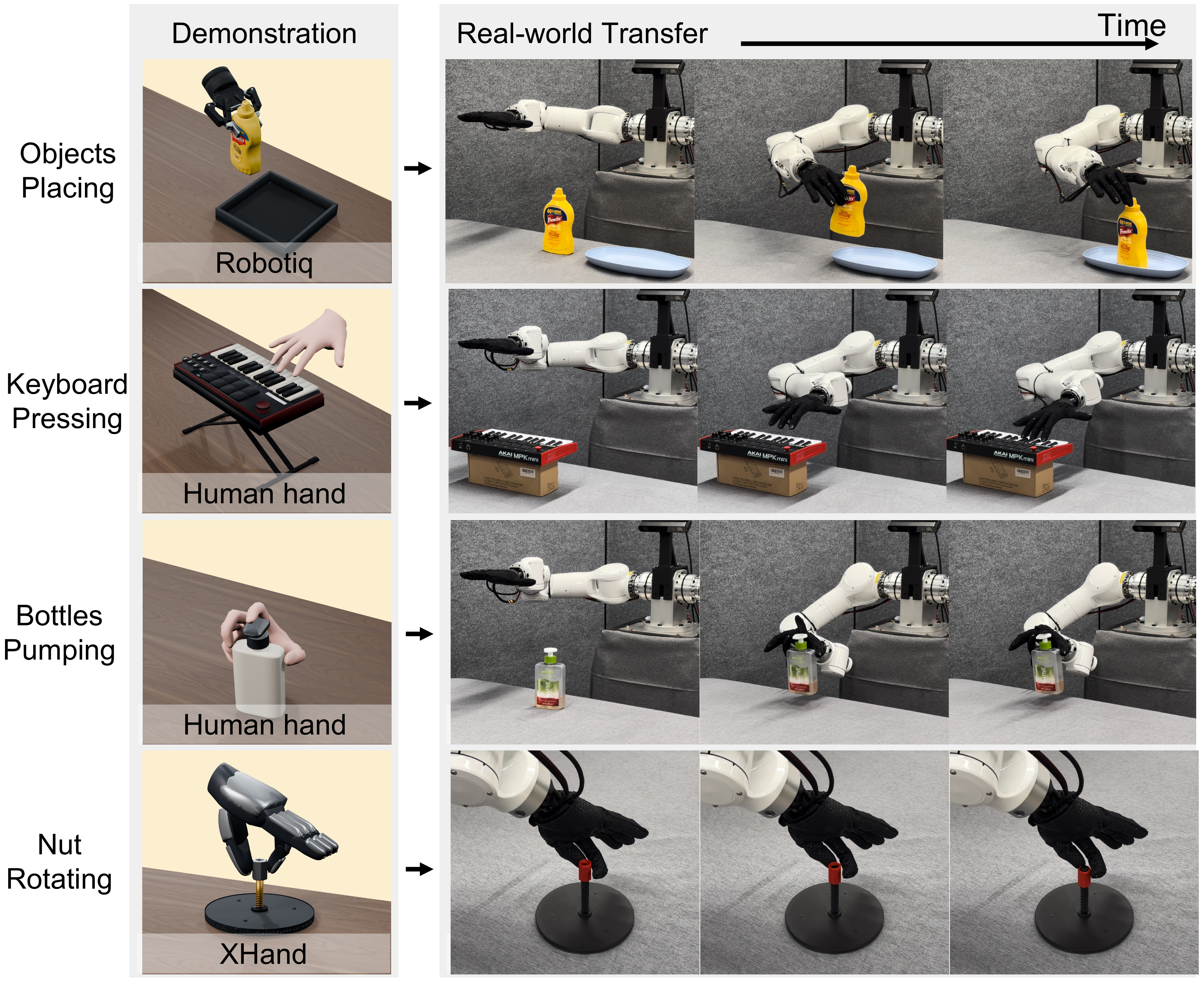}\par
    \caption{\textbf{Real-world transfer.} Source demonstrations (left) and zero-shot Wuji execution (right) across four tasks.\strut}
    \label{fig:real-world-rollouts}
    \vspace{-2pt}
\end{figure}

We evaluate all four tasks on a Wuji hand mounted on a Tianji Marvin arm in the real world, using the same frozen kinematic decoder without target-task demonstrations or real-world fine-tuning, as shown in \Cref{fig:real-world-rollouts}. The task settings follow simulation, with real-world objects shown in \Cref{fig:experiment-scope}c. Objects placing and bottles pumping each use one object with geometry similar to one of the simulation instances. The demonstration sources are Robotiq for objects placing, a human hand for keyboard pressing and bottles pumping, and XHand for nut rotating. For each human-source task, one recorded demonstration is augmented to $800$ episodes by randomizing object positions and interpolating the approach motion, with observations rendered through kinematic replay in simulation. The human hand is represented with MANO~\citep{MANO:SIGGRAPHASIA:2017}, allowing its demonstrations to be encoded into the same latent intent space.

The simulation workspace matches the hardware setup. A calibrated, base-mounted ZED camera reconstructs the scene point cloud. We retain object points using SAM~2~\citep{ravi2025sam2} segmentation masks and remove spurious depth points before passing the point cloud to the interaction policy.
Gaussian noise is added to the training observations to reduce the perception gap. We evaluate transfer from both robot and human demonstrations over $10$ trials per task, reporting success rates for all four tasks in \Cref{tab:real-world-results}. For nut rotating, success requires rotating the nut by more than $\pi$ radians. We use this binary criterion because precise rotation angles are difficult to measure reliably in the real world.

\section{Limitations}
\label{sec:limitations}

\MethodAcronym\ assumes that manipulation tasks can be described as sequences of contact patterns between the manipulator and the object.
Tasks where contact is incidental to the success criterion, or where the target embodiment cannot realize the source contact pattern, fall outside its scope.
The framework transfers source-discovered strategies through contact rather than inventing new strategies tailored to the target morphology.
In addition, the kinematic decoder is trained from the kinematic model alone, so compliant pads, tendon slack, and cable coupling introduce a sim-to-real gap that this supervision does not address.

\section{Conclusion}

We presented \MethodAcronym, a framework for zero-shot cross-embodiment manipulation that decouples interaction prediction from embodiment-specific kinematic decoding through contact-based interaction intents.
The interaction policy learns from source demonstrations, while the kinematic decoder learns to realize these intents from kinematic models alone.
Across four tasks and six robot embodiments in simulation, \MethodAcronym transfers without target-task demonstrations and achieves higher average success than the evaluated embodiment-conditioned and trajectory retargeting baselines on their supported settings. 
Real-world deployment further shows transfer from robot and human demonstrations to a Wuji hand.
More broadly, our results suggest that representing actions as intended interactions makes task knowledge more reusable across robots, while each robot learns how to realize those interactions from its kinematic model.


\section*{ACKNOWLEDGMENT}
We gratefully acknowledge use of the research computing resources of the Empire AI Consortium, Inc~\cite{bloom2025empire}, with support from Empire State Development of the State of New York, the Simons Foundation, and the Secunda Family Foundation. This work was supported in part by the Amazon Research Awards and an NVIDIA Academic Grant. We thank Yunhao Cao, Mengling Zhang, Jialiang Zhang, Chuanruo Ning, Xingyi He, Adhitya Polavaram, Zhenyu Wei, Qi Wu, Cory Fan, and Pranav Thakkar for constructive discussions and feedback. We thank Calvin Qiu, Adhitya Polavaram, Yunhao Cao, and Cory Fan for their generous help on real world and simulation robot infrastructure. 


\bibliographystyle{IEEEtran}
\bibliography{main}  
\clearpage
\providecommand{\mainref}[1]{\ref{#1}}
\providecommand{\maineqref}[1]{\eqref{#1}}
\providecommand{\mainCref}[1]{\Cref{#1}}
\newcommand{\appendixtodo}[2]{%
    \par\begingroup\small\raggedright\color{blue}%
    \noindent\textbf{[TODO #1:} #2\textbf{]}\par\endgroup}
\newcommand{\appendixpending}{\textcolor{blue}{TODO}}
\newcommand{\codeadded}[1]{{\color{blue}#1}}

\useRomanappendicesfalse
\appendices
\raggedbottom
\pagestyle{plain}
\hypersetup{hidelinks}
\setcounter{equation}{0}
\renewcommand{\theequation}{A\arabic{equation}}
\renewcommand{\theHequation}{appendix.\arabic{equation}}
\setcounter{table}{0}
\renewcommand{\thetable}{A\arabic{table}}
\renewcommand{\theHtable}{appendix.\arabic{table}}
\setcounter{figure}{0}
\renewcommand{\thefigure}{A\arabic{figure}}
\renewcommand{\theHfigure}{appendix.\arabic{figure}}

\section{Method Implementation Details}
\label{app:implementation-details}

We provide the implementation details of the intent encoder, kinematic decoder, and interaction policy. Following \mainCref{sec:method-kd,sec:method-ip}, we first train the intent encoder and decoder bank on synthetic kinematic data, then train each interaction policy on source demonstrations preprocessed into latent intents.

\subsection{Synthetic Kinematic Data}
\label{app:decoder-details}
\label{app:decoder-data-synthesis}

\noindent\textbf{Surface contact template.}
For each embodiment $e$, we precompute a surface contact template $\mathcal{T}^{e}$ from its kinematic model $\beta^{e}$. We use $4096$ surface points for every robot embodiment. Each robot surface point stores a position $\bar{x}_r$ and an outward normal $\bar{n}_r$ in the frame of its parent link $\ell_r$. Forward kinematics maps these points and normals into the world frame at any configuration. For MANO, we use all $778$ vertices of the original model and obtain their positions and normals from the posed hand mesh.

\noindent\textbf{Sampling interaction intents.}
We sample a contact configuration $q_{\mathrm{contact}}$ within the joint limits. For embodiments with a movable base, the base translation is sampled uniformly in $[-0.40,0.40]^3$\,m and the base rotation uniformly from $\mathrm{SO}(3)$. We draw the number of contact samples $K_{\mathrm{syn}}$ uniformly from $\{2,\ldots,6\}$. For each contact, we first select a nonempty finger or palm region uniformly, then select a template point uniformly within that region. Contacts are sampled independently with replacement. This gives each region equal sampling probability regardless of its number of template points. Let $S$ contain the selected template indices. The resulting interaction intent is
\begin{equation}
    c = \bigl\{(x_r(q_{\mathrm{contact}}),\,
        n_r(q_{\mathrm{contact}}))\bigr\}_{r\in S},
    \label{eq:app-synthetic-intent}
\end{equation}
where $x_r(q)$ and $n_r(q)$ denote the world-frame position and normal of template point $r$ under forward kinematics. This construction uses only the embodiment's kinematic model, without an object model or task demonstration.

We obtain $q_{\mathrm{init}}$ by perturbing $q_{\mathrm{contact}}$ with zero-mean Gaussian noise. The standard deviations are $5\%$ of each joint's range, $0.005$\,m for base translation, and $5^\circ$ for base rotation. Joint values are then clamped to their limits.

For Robotiq+Stick, the composite sampler draws contacts on one or both independently actuated components, with one to three contacts per selected component. The gripper and stick have separate base poses, sampled within $0.20$\,m of each other.

\noindent\textbf{Supervision of contact-relevant joints.}
The joint mask $m\in\{0,1\}^{d_e}$ identifies the joints that affect the selected points. We set $m_j=1$ if joint $j$ lies upstream of a selected point in the kinematic tree, and $m_j=0$ otherwise. The goal configuration follows Eq.~\maineqref{eq:goal-config}: contact-relevant joints take their values from $q_{\mathrm{contact}}$, while the remaining joints retain their values from $q_{\mathrm{init}}$. Each example in $\mathcal{D}_{g}^{e}$ is therefore a triple $(c,q_{\mathrm{init}},q_{\mathrm{goal}})$.

\subsection{Intent Encoder and Kinematic Decoder}
\label{app:decoder-training-details}

\begin{figure}[t]
    \centering
    \includegraphics[width=\linewidth]{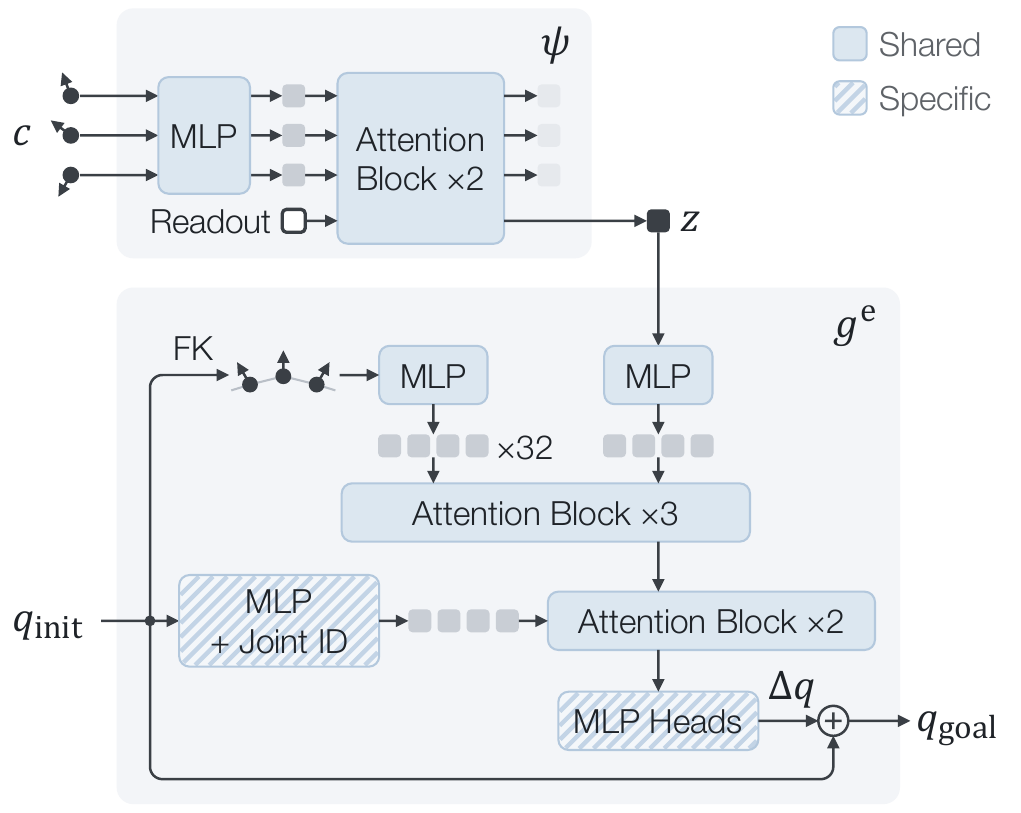}
    \caption{\textbf{Intent encoder and kinematic decoder architecture.} The encoder $\psi$ maps the interaction intent $c$ to the latent intent $z$. Conditioned on $z$ and $q_{\mathrm{init}}$, the decoder $g^{e}$ predicts a residual $\Delta q$ that is composed with $q_{\mathrm{init}}$ to obtain $q_{\mathrm{goal}}$. Small squares denote tokens, FK denotes forward kinematics, and $\times N$ indicates $N$ stacked attention blocks. As in \Cref{fig:method-overview}, solid fills indicate shared modules and hatching indicates embodiment-specific modules.}
    \label{fig:app-architecture}
\end{figure}

\noindent\textbf{Architecture.}
\Cref{fig:app-architecture} shows the intent encoder $\psi$ and kinematic decoder $g^{e}$. Since the interaction intent $c$ is an unordered set, $\psi$ must be invariant to the order. An MLP projects each contact position and normal to a $256$-dimensional token, and two attention blocks process the contact tokens. A learned readout token aggregates their features into the latent intent $z\in\mathbb{R}^{256}$. We mask padded contacts and omit sequence-order encodings to preserve permutation invariance. Sharing $\psi$ across embodiments encourages the latent intent space to carry consistent meaning across them.

The kinematic decoder follows the heuristic of matching intended contacts to suitable parts of the embodiment through attention between intent and surface tokens. Before training, we select $32$ points by farthest-point sampling on each embodiment's canonical surface template and keep their indices fixed. Given $q_{\mathrm{init}}$, forward kinematics (FK) maps these points and their outward normals to the world frame, and an MLP projects them into surface tokens. A second MLP expands $z$ into four intent tokens, each with $256$ dimensions, matching the surface tokens. Three attention blocks exchange information between the two sets of tokens, allowing the intended contacts to be related to the embodiment's surface at its initial configuration.

To predict the configuration change that realizes the intent, embodiment-specific MLPs project $q_{\mathrm{init}}$ into query tokens, with joint-identity embeddings (Joint ID in \Cref{fig:app-architecture}) distinguishing the joints. Two attention blocks let these query tokens attend to the updated surface tokens. Embodiment-specific MLP heads then map the resulting features to the configuration residual $\Delta q$. Translation and joint residuals are added to their values in $q_{\mathrm{init}}$, while the base-rotation residual uses the first two columns of a rotation matrix as six continuous coordinates and is composed with the initial rotation. Together, these updates give $q_{\mathrm{goal}}$.

To improve parameter efficiency and learn common kinematic features across embodiments, the decoders share the MLPs that produce surface and intent tokens and all attention blocks. Only the configuration MLPs, joint-identity embeddings, and output MLP heads are embodiment-specific, as indicated by the hatched modules in \Cref{fig:app-architecture}.

\noindent\textbf{Training.}
We jointly train $\psi$ and the decoder bank $\{g^{e}\}$ with the regression loss in Eq.~\maineqref{eq:low-loss}. Training covers all embodiments and runs for $500$ epochs with batch size $128$ per embodiment. The implementation averages the per-embodiment losses equally. For translation and joint residuals, we apply the same per-coordinate scale factors as for absolute configurations, without an offset, so zero displacement remains zero. For the six rotation coordinates, we subtract the identity-rotation vector $(1,0,0,0,1,0)$, so no rotation also maps to zero. We synthesize $10^6$ training examples and $1000$ validation examples per embodiment. Training uses AdamW with learning rate $10^{-4}$, $(\beta_1,\beta_2)=(0.9,0.999)$, weight decay $10^{-6}$, and gradient clipping at norm $1$. The learning-rate schedule uses $2000$ warmup steps followed by cosine decay. The trained encoder and decoder bank are frozen during interaction-policy training and reused across tasks.

\subsection{Preprocessing Source Demonstrations}
\label{app:policy-data-preprocessing}

\noindent\textbf{Contact scoring.}
We use a distance- and normal-based score adapted from GenDexGrasp~\citep{li2023gendexgrasp} to identify the source surface regions involved in each demonstration. At time $t$, let $O_t$ denote the object point cloud used for contact extraction, and let $p_{t,r}$ and $n_{t,r}$ denote the world-frame position and outward normal of source template point $r$. For an object point $y\in O_t$, define
\begin{equation}
    \gamma_{t,r,y} =
    \frac{(y-p_{t,r})^{\top}n_{t,r}}
         {\|y-p_{t,r}\|_2+\epsilon},
    \label{eq:app-contact-alignment}
\end{equation}
where $\epsilon=10^{-5}$ prevents division by zero. The aligned distance and contact score are
\begin{equation}
    \begin{aligned}
        d_{t,r} &= \sqrt{\min_{y\in O_t}
            \left[\|p_{t,r}-y\|_2
            \exp\!\left(2(1-\gamma_{t,r,y})\right)\right]},\\
        s_{t,r} &= 2\left(1-\sigma(10d_{t,r})\right),
    \end{aligned}
    \label{eq:app-contact-score}
\end{equation}
with $\sigma$ denoting the sigmoid function. A template point becomes a contact candidate when $s_{t,r}\geq 0.4$.

\noindent\textbf{Trajectory-level point selection.}
We pool the template indices whose scores exceed the threshold in at least one frame. Following \mainCref{sec:method-ip}, we determine the number of clusters from the average number of connected contact regions per frame, capped at six, while retaining at least two indices overall. We cluster the candidate positions in the canonical template frame with $k$-means. Within each cluster, we sum each candidate's above-threshold scores over the trajectory and retain the index with the largest sum. This favors surface points that consistently participate in the demonstrated interaction.

Forward kinematics then maps the selected indices to world-frame positions and normals at every time step, forming $c_t$. We retain these points during approach and withdrawal, including frames without active contact. Thus, the interaction intent describes the motion of the surface regions involved in the task throughout the demonstration.

\noindent\textbf{Latent intent labels.}
The frozen intent encoder maps each $c_t$ to $z_t=\psi(c_t)$, giving training triples $(o_t,z_{t-1},z_t)$. Source configurations are used to extract the interaction intents, but are not inputs to the interaction policy.

\subsection{Interaction Policy}
\label{app:policy-details}
\label{app:policy-training-details}

The iDP3 interaction policy uses two observation steps and a prediction horizon of $16$, returning $15$ latent intents per prediction. An XYZ-only PointNet produces a $128$-dimensional observation feature. The diffusion network uses FiLM conditioning, channel widths $(256,512,1024)$, and a $128$-dimensional diffusion-step embedding. We use $50$ noise steps with a squared-cosine schedule and $10$ DDIM steps at inference, predicting the clean action sample.

Training runs for $200$ epochs with batch size $64$, using AdamW with learning rate $10^{-4}$, $(\beta_1,\beta_2)=(0.95,0.999)$, weight decay $10^{-6}$, and gradient clipping at norm $1$. After $500$ warmup steps, the learning rate remains constant. We select the exponential moving average (EMA) checkpoint with the lowest validation loss.

At the start of each training trajectory, we set $z_{-1}=z_0$, the first encoded intent of that source demonstration. At deployment, we initialize the previous latent intent with the mean of the first encoded intents over the training episodes. We fill both observation-history slots with the initial point cloud and this latent intent. Thereafter, each executed intent becomes the previous-intent input to the policy.

\section{Simulation Experiment Details}
\label{app:experiment-details}

\subsection{Observations and Control}
\label{app:simulation-setting-details}

In MuJoCo, arm-mounted embodiments use inverse kinematics to convert wrist-pose commands to arm joint targets; floating embodiments command their base poses directly. A position controller tracks the configuration targets.

Simulator segmentation and workspace filtering remove robot and table points. We store $4096$ object points per frame and downsample to $256$ XYZ points for the policy encoder. Action channels are normalized using training-data bounds. Objects placing, keyboard pressing, and bottles pumping use $120$-frame rollouts at a nominal $20$\,Hz.

\subsection{Task Assets and Settings}
\label{app:task-details}
\label{app:evaluation}

\noindent\textbf{Object assets.}
Objects placing uses six YCB objects: pudding box, bowl, potted meat can, mug, sugar box, and mustard bottle. Bottles pumping uses two assets, SoapDispenser002 and SoapDispenser007, from NVIDIA's PhysicalAI kitchen-object dataset,\footnote{\href{https://huggingface.co/datasets/nvidia/PhysicalAI-Robotics-Manipulation-Objects-Kitchen-MJCF}{NVIDIA PhysicalAI Robotics Manipulation Objects Kitchen (MJCF).}} together with a third pump bottle reconstructed from our own scan, as shown in \mainCref{fig:experiment-scope}b.

\noindent\textbf{Nut-rotating initial poses.}
We construct one initial pose per target embodiment from the recorded XHand reset pose. We first align the palm position, the palm-to-contact direction, and the direction between the two pinch contacts. We then refine the base translation and participating finger joints to match the contact positions and normals at the source's initial pose. Refinement respects joint limits and object clearance while keeping the aligned base orientation fixed. The resulting pose is reused across test episodes.

\noindent\textbf{Evaluation.}
We use training seeds $1$, $2$, and $3$ with data-split seed $42$; GET-Zero uses training seeds $42$, $43$, and $44$. The keyboard dataset reserves $100$ of $900$ source episodes for evaluation. For the nut-rotation metric in \mainCref{sec:exp-setup}, we measure signed hinge displacement. Source displacement is the final angle minus the initial angle; target displacement is measured from a zero-angle reset.

\subsection{Baseline Implementations}
\label{app:baseline-setting-details}

We describe the components retained from each method and the adaptations to our tasks. Task supervision comes only from source demonstrations; target kinematic models provide the embodiment structure needed for transfer.

\noindent\textbf{GET-Zero.}
We retain GET-Zero's graph-conditioned Transformer, shared joint-action and forward-kinematics heads, and six-step history of measured joint states and previous commands~\citep{patel2025getzero}. To incorporate scene observations, we add a PointNet that maps $256$ object points to a $128$-dimensional feature. A shared wrist-pose head enables the translational and rotational motion required by the tabletop tasks. Joint features include native joint limits, axes, and tendon coupling to represent the different hand and gripper structures. Objects placing, keyboard pressing, and bottles pumping use incremental wrist commands. Nut rotating retains fixed-base control.

We train by behavior cloning on the source demonstrations, retaining the auxiliary forward-kinematics loss with weight $2$ and computing its labels only at source demonstration states. We follow the official implementation's Adam optimizer and learning-rate schedule: a peak rate of $1.5\times10^{-4}$, linear warmup over $15\%$ of updates, and inverse-square-root decay. We adapt the training budget to our source-only task datasets, using $100$ epochs with batch size $256$. To select a policy for closed-loop execution, we evaluate checkpoints at epochs $10$, $25$, $50$, $75$, and $100$, together with the checkpoint having the lowest source validation action loss, on $24$ fixed source episodes. We select by rollout score, using source action loss to break ties.

\noindent\textbf{OHRA.}
We use OHRA's extended canonical representation and its zero-shot grasp-generation implementation, which conditions directly on $152$ hand parameters without a parameter VAE~\citep{wei2026onehand}. We retain its $100$-epoch training schedule, batch size $128$, and initial learning rate $10^{-3}$, with $1000$ diffusion noise steps and $10$ DDIM steps at inference. For keyboard pressing, we replace grasp supervision with one canonical pre-press pose per source episode, conditioned on the initial scene point cloud. The predicted pose is mapped to the target hand and executed with a scripted approach and press lasting $96$ and $24$ frames, respectively.

\noindent\textbf{D(R,O).}
We retain D(R,O)'s contrastive robot-encoder pretraining, distance-matrix prediction, and configuration recovery through multilateration and link-pose alignment~\citep{wei2024drograsp}. To pretrain without target-task demonstrations, we replace grasp-derived configuration pairs with $900$ synthetic perturbation pairs for each of Robotiq, Wuji, Allegro, and Barrett. We follow the official implementation's training settings: $100$ epochs for robot-encoder pretraining, followed by $200$ epochs of task training with the encoder frozen, batch size $8$, and learning rate $10^{-4}$. Penetration loss is retained for objects placing but disabled for keyboard pressing because a static surface model cannot represent key motion.

Both D(R,O) and OHRA use the available $256$-point initial observations in place of the original $512$-point inputs, retaining the original $0.002$\,m Gaussian noise during training. Ground-truth object poses are used for object-frame canonicalization. D(R,O) predicts a grasp or pre-press pose from this initial observation, followed by $64$ configuration-recovery iterations. For objects placing, we retain its grasp-closing controller and add scripted approach, lift, transport, and release, with a $0.10$\,m approach offset and lift. Keyboard pressing uses the same $96$-frame approach and $24$-frame press as OHRA.

\noindent\textbf{SPIDER.}
We follow SPIDER's trajectory-retargeting procedure, using kinematic initialization followed by sampling-based optimization through MuJoCo-Warp physics rollouts~\citep{pan2025spider}. The optimizer uses $1024$ samples per iteration, up to $16$ iterations, a $0.4$\,s horizon, $0.1$\,s control and knot intervals, and a $0.01$\,s planning step. We provide source embodiment and object trajectories and manually specify task-dependent part correspondences. For keyboard pressing, Wuji index, middle, and ring fingers map to the same Allegro fingers or to Barrett middle, thumb, and index fingers. For bottles pumping, Wuji thumb, index, and middle fingers map to the corresponding Allegro fingers or to Barrett middle, thumb, and index fingers. Nut rotating maps source index and thumb contacts to the corresponding target anchors and replaces free-object motion with the nut's hinge angle. The optimized commands are replayed with the target's task controller.

\subsection{Ablation Implementation}
\label{app:ablation-details}

\noindent\textbf{Raw contact slots.}
The w/o Latent policy predicts contact positions and normals in six fixed slots, giving a $36$-dimensional action. The policy architecture and training settings otherwise follow the latent-intent policy. Unused slots are zero-padded and masked in the intent encoder. Position and normal channels use training-data-based action normalization. At execution, we undo action normalization and pass the valid contacts through the frozen intent encoder and kinematic decoder. Padded slots are set to zero before execution and before feeding the previous prediction back to the policy.

\noindent\textbf{Initial-pose perturbations.}
For \mainCref{tab:initial-pose-robustness}, translations are along world $+z$ and rotations are about the hand's local $x$ axis. The small and large conditions combine translation and rotation; the flip changes only orientation, and the shift changes only position. Initial finger joint angles remain unchanged. Each condition uses $100$ episodes.

\section{Real-World Deployment Details}
\label{app:real-world}

We deploy all four tasks on a Wuji hand mounted on a Tianji Marvin arm. As in \mainCref{sec:exp-real}, execution uses the frozen kinematic decoder without target-task demonstrations or real-world fine-tuning.

\subsection{Source Demonstrations}
\label{app:real-world-demonstrations}

The source embodiments are Robotiq for objects placing, a human hand for keyboard pressing and bottles pumping, and XHand for nut rotating. For each human-source task, we record one demonstration and augment it to $800$ episodes by randomizing object positions and interpolating the approach motion. Observations are rendered through kinematic replay in simulation. The human hand is represented with MANO~\citep{MANO:SIGGRAPHASIA:2017}, whose surface geometry and configuration-to-surface mapping allow us to extract interaction intents and encode them in the same latent space.

\balance
\subsection{Perception}
\label{app:real-world-setup}

The simulation workspace matches the hardware setup. A calibrated, base-mounted ZED camera reconstructs the scene point cloud. SAM~2~\citep{ravi2025sam2} masks retain the task objects, followed by workspace and depth filtering. We capture $1280\times720$ images at a requested $15$\,Hz and retain depths between $0.15$ and $2.0$\,m. Camera points are transformed into the common robot-base frame using hand--eye calibration. Objects placing and bottles pumping each use one real object with geometry similar to a simulation instance.

We augment the policy's training observations with Gaussian point-cloud noise to reduce the perception gap. For objects placing, the noise standard deviation is $0.004$\,m, without point dropout. The intent encoder and kinematic decoder remain frozen.

\subsection{Control and Evaluation}
\label{app:real-world-protocol}
\label{app:real-world-tasks}

The interaction policy predicts latent intents from the live object-only point cloud and the previous latent intent. The Wuji kinematic decoder maps each executed intent and the current configuration $q_t^{\mathrm{Wuji}}$ to a commanded next configuration, following Eq.~\maineqref{eq:deploy}. The arm controller realizes the wrist motion, while the hand controller tracks the joint targets.

The hardware profile requests a $5$\,Hz policy execution rate and a $90$\,Hz arm-command stream. Wrist poses are converted through inverse kinematics, and intermediate commands interpolate between consecutive targets. Perception and inverse-kinematics overruns lengthen the rollout rather than skip intermediate targets. For nut rotating, the arm holds a fixed pose while the hand executes joint commands. For keyboard pressing, the keyboard rests on a $0.10$\,m-high support box to prevent collisions between the arm and the table.

We run $10$ trials per task. Objects placing, keyboard pressing, and bottles pumping use the task criteria in \mainCref{sec:exp-setup}. For real-world nut rotating, success requires more than $\pi$ radians of rotation. This binary criterion differs from the mean clipped rotation ratio used in simulation because precise rotation angles are difficult to measure reliably on hardware.

\end{document}